\documentclass{article}

\PassOptionsToPackage{numbers, compress}{natbib}
\usepackage[preprint]{neurips_2026}
\usepackage{graphicx}
\usepackage[utf8]{inputenc} % allow utf-8 input
\usepackage[T1]{fontenc}    % use 8-bit T1 fonts
\usepackage{amsmath}
\usepackage{hyperref}       % hyperlinks
\usepackage[capitalize, noabbrev]{cleveref}
\usepackage{url}            % simple URL typesetting
\usepackage{booktabs}       % professional-quality tables
\usepackage{subcaption}
\usepackage{amsfonts}       % blackboard math symbols
\usepackage{amssymb}
\usepackage{nicefrac}       % compact symbols for 1/2, etc.
\usepackage{microtype}      % microtypography
\usepackage[dvipsnames]{xcolor}
\usepackage{colortbl}
\usepackage{enumitem}
\usepackage{wrapfig}
\newcommand{\second}[1]{#1}
\newcommand{\maketitlesupplementary}{\section*{Supplementary Material}}
\title{Two Global Crops Suffice: Locating Semantic Emergence in DINO-Style Self-Supervised Learning}

\author{%
  Basavaraj Sunagad \\
  CISPA Helmholtz Center for Information Security \\
  \And
  Artur Jesslen \\
  University of Freiburg \\
  \And
  Adam Kortylewski \\
  CISPA Helmholtz Center for Information Security \\
}

\begin{document}

\maketitle

\begin{abstract}
Self-supervised vision transformers trained with DINO-style objectives exhibit striking emergent semantic representation quality across visual tasks, yet the mechanisms underlying this behavior remain unclear. We present a systematic empirical dissection of the DINO family and show that semantic representations arise primarily from enforcing consistency between geometrically distinct global views of the same image instance. This instance-specific global alignment acts as the semantic anchor of DINO-style learning. Across controlled retraining experiments evaluated on semantic correspondence and a diverse suite of 2D and 3D downstream tasks, we find that patch-level masking objectives enhance semantics only when trained jointly with this global alignment, indicating that the iBOT objective refines and densifies existing semantic structure rather than creating it independently. In contrast, local-to-global view alignment does not substantially improve semantic qualities at fixed compute beyond a purely global alignment. Beyond training design, we revisit how semantic representation quality should be evaluated: while classification accuracy is the standard validation score, semantic correspondence provides a complementary axis that more reliably predicts downstream task performance. Together, these findings provide a functional decomposition of DINO-style learning and represent an important step toward understanding how semantic representations emerge in self-supervised vision models.
\end{abstract}

\begin{figure}[t]
    \centering
    \includegraphics[width=\linewidth]{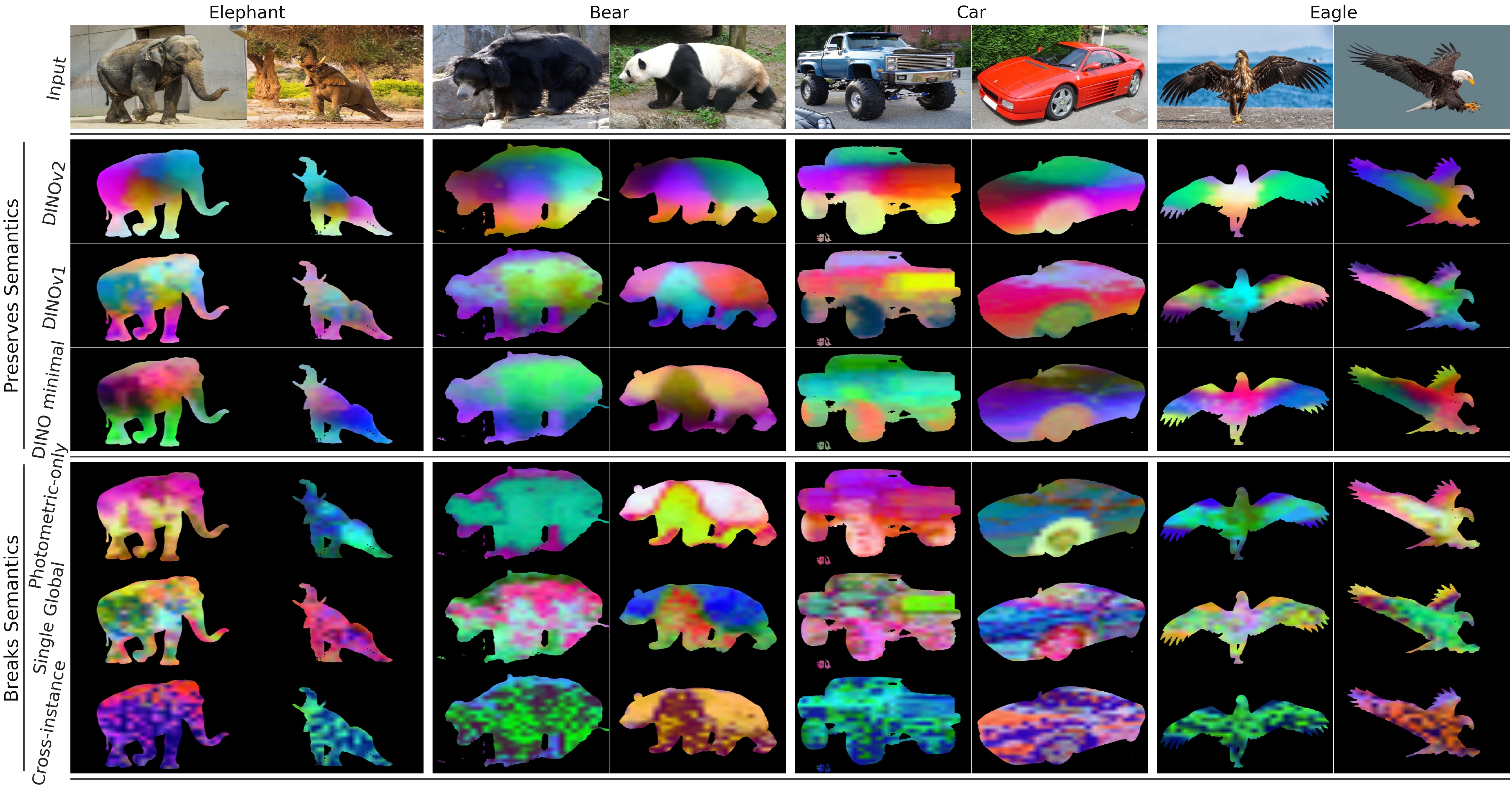}
    \caption{%
      \textbf{PCA visualization of patch features under different training objectives.}
Each row shows RGB maps of the first three principal components (PC1--3) of ViT-L/16
patch features, fitted jointly per category over masked foreground tokens (backgrounds set to black).
\textit{Preserves Semantics (top).}
\textit{DINO minimal} (2 global crops of the same image) already suffices for part-consistent structure to emerge.
\textit{DINOv1} (2G+8L, DINO loss only, PCK@.1\,=\,32.0) closely matches \textit{DINO minimal} (31.2),
confirming local crops are not the source of semantic structure.
\textit{DINOv2} adds iBOT to the same recipe and yields the richest part structure (42.5).
\textit{Breaks Semantics (bottom).}
\textit{Photometric-only} (1 global crop with photometric augmentation only, PCK\,=\,20.1)
produces weaker, patchier gradients.
\textit{Single Global} (1 global\,+\,8 local crops, PCK\,=\,10.6) and
\textit{Cross-instance} (views from two different images, PCK\,=\,7.3)
both collapse part-consistent structure into fragmented, mottled patterns.
}
    \label{fig:teaser}
\end{figure}
\section{Introduction}
\label{sec:intro}
Self-supervised vision transformers trained with DINO-style~\cite{caron2021emerging,zhou2021ibot,oquab2023dinov2} objectives have emerged as strong general-purpose visual backbones, exhibiting striking semantic structure: features cluster by category, attention maps delineate object boundaries, and learned representations transfer broadly across dense prediction tasks~\cite{oquab2023dinov2}. The standard training recipe is a tightly coupled system, developed through large-scale empirical engineering: it combines multi-crop view generation~\cite{caron2020swav}, patch-level masked prediction~\cite{zhou2021ibot}, careful augmentation design~\cite{dosovitskiy2014discriminative}, and momentum distillation~\cite{grill2020bootstrap}. Each component is presented as beneficial, but their individual contributions to semantic representation quality, as opposed to classification accuracy, have not been cleanly isolated. This is the gap our work addresses.

The central ambiguity concerns the interplay between three components of the recipe: global-view contrast between two large crops of the same image, local-to-global alignment that pairs small local crops with global views, and patch-level masked prediction that operates at the token level. Patch-level supervision could plausibly provide the dense correspondence signal underlying semantic structure; equally, global-global agreement alone may constitute a sufficient semantic training signal, with local and patch objectives serving auxiliary roles. The question is not whether these components improve overall benchmark numbers (they do), but which of them creates the semantic structure that DINO-family models are valued for, versus which refine an already-semantic representation.

This question fits into a broader research thrust on identifying minimal principles for effective self-supervised learning. Methods such as SimDINO~\cite{wu2025simplifying} and LeJEPA~\cite{balestriero2025lejepa} have shown that many components in the DINOv2 recipe exist primarily to prevent representational collapse, and can be replaced by explicit regularization without loss of performance. Our work is complementary but diagnostic: rather than proposing new objectives, we ask which existing components are responsible for semantic emergence in the learned representation, and which primarily refine downstream task performance such as classification.
We conduct controlled pretraining experiments at a moderate scale, systematically ablating crop configuration (global-only vs. global+local), auxiliary objectives (with and without patch-level prediction), and crop scale. We evaluate all variants using a comprehensive suite that includes classification (kNN, linear probe), semantic and geometric correspondence (SPair-71k, NAVI), and dense downstream tasks spanning depth estimation, pose, point tracking, and semantic segmentation (NYU, ImageNet3D, TAPVid, ADE20K). This dual evaluation axis, classification alongside correspondence, turns out to be essential for correctly interpreting representation quality.

Our central finding is that same-instance global alignment, consistency between two geometrically displaced global crops of the same image, acts as a semantic anchor for DINO-style learning. Two global crops with the DINO loss alone, without local crops or patch-level prediction, already yield representations with strong semantic structure and competitive performance across the dense evaluation suite. Adding local crops improves classification accuracy substantially but leaves correspondence largely unchanged; adding patch-level prediction (iBOT) refines correspondence quality (~10.5 PCK points) but is not its source. Necessity tests further confirm that removing geometric displacement between views, or replacing same-image crops with crops from different instances, each collapses semantic correspondence, pinpointing same-instance geometric alignment as the operative inductive bias.

This observation motivates a closer look at how representation quality is measured. Classification accuracy is sensitive to recipe choices in ways that do not predict dense downstream utility: in fixed-scale crop experiments, classification peaks at s=0.40 while correspondence and dense downstream tasks peak at s=0.60–0.72. Across our experimental sweep, semantic and geometric correspondence correlate substantially more strongly with dense downstream performance than kNN or linear probes. We therefore argue that semantic correspondence should be treated as a first-class evaluation axis alongside classification, not as a replacement, but as a necessary complement when the goal is to understand semantic emergence rather than classification transfer.

Through a systematic and principled decomposition of the DINO training recipe, we arrive at the following findings:
\begin{enumerate}[label=\textit{(\roman*)}, leftmargin=*, itemsep=0.25em]
\item \emph{Same-instance global-view alignment is the semantic anchor of DINO-style SSL}: it is sufficient to produce strong semantic patch features, and removing geometric displacement or same-instance identity each collapses semantic correspondence. Local crops and patch-level prediction refine this representation (local crops by contributing to classification accuracy through view diversity, iBOT by sharpening correspondence quality), but neither creates the semantic structure.
\item \emph{Validating SSL recipes by classification accuracy can systematically misjudge dense-task utility}: classification and dense correspondence have opposing crop-scale optima (s=0.40 vs. s=0.60–0.72 in fixed-scale experiments).
\item \emph{Semantic and geometric correspondence predict dense downstream performance more reliably than kNN or linear probes} across our experimental sweep, motivating their inclusion as first-class evaluation axes for SSL.
\end{enumerate}
\section{Related Work}
\label{sec:rel_work}
Self-supervised visual representation learning has advanced through increasingly complex training recipes, yet the mechanisms responsible for semantic structure remain poorly understood.
We review the lines of work most relevant to this diagnostic question, spanning view generation strategies, training objectives, and evaluation methodology.

\textbf{Self-supervised learning and the two-view paradigm.}
Self-supervised visual representation learning has converged on a two-view paradigm: a network produces similar embeddings for augmented views of the same image and dissimilar ones for different images~\cite{dosovitskiy2014discriminative,chen2020simclr,he2020moco}. Subsequent methods showed that competitive representations emerge from two-view agreement even without negative pairs~\cite{grill2020bootstrap,chen2021simsiam}, with redundancy-reduction methods~\cite{zbontar2021barlow,bardes2022vicreg} offering an alternative route to collapse avoidance via explicit feature decorrelation. We take this paradigm as our starting point and identify \textit{which specific properties} of two-view alignment (geometric displacement across views, same-instance identity, and the global contrastive loss) are each necessary for semantic structure to emerge.

\textbf{Multi-crop and local-global view strategy.}
SwAV~\cite{caron2020swav} introduced the multi-crop strategy, mixing two large global crops with several small local crops per image. DINO~\cite{caron2021emerging} adopted this within a self-distillation framework where local views match representations derived from global views, and multi-crop has since become a de-facto standard in DINO-family recipes. Whether local-to-global alignment specifically contributes to \textit{semantic} representation quality, as opposed to improving classification or training efficiency, has not been systematically isolated. Our ablations show that removing all local crops leaves correspondence largely unchanged, challenging the assumption that local-to-global matching is the operative mechanism for semantic emergence.

\textbf{Patch-level objectives and masked image modeling.}
Masked image modeling~\cite{bao2022beit,he2022mae} strengthens spatial representations through token-level supervision; iBOT~\cite{zhou2021ibot} integrated this into the DINO self-distillation framework, and DINOv2~\cite{oquab2023dinov2} adopted it alongside large-scale data curation. Token-level supervision is widely interpreted as the primary driver of dense semantic correspondence. We directly test this assumption: removing iBOT degrades but does not collapse correspondence, while training with iBOT alone produces near-zero correspondence, establishing it as a \textit{refiner} rather than a \textit{creator} of semantic structure.

\textbf{Simplifying self-supervised learning pipelines.}
A growing body of work simplifies SSL recipes by identifying which components are necessary for stable optimization. SimDINO~\cite{wu2025simplifying} replaces DINOv2 heuristics with explicit variance regularization without performance loss; LeJEPA~\cite{balestriero2025lejepa} proposes more principled objective formulations; recent work also shows that augmentation needs decrease with data scale~\cite{moutakanni2024augmentation}. These efforts ask which components can be removed without accuracy loss; we ask which components \textit{create} semantic structure: orthogonal questions, since a component can be removable (replaced by a functionally equivalent alternative) yet still be the causal origin of a capability.

\textbf{Probing learned representations.}
Evaluating SSL representations has historically relied on classification-based probes~\cite{chen2020simclr,caron2021emerging}, which provide only an indirect view of the spatial structure in dense patch representations. Recent work uses ViT patch features as zero-shot dense descriptors for semantic correspondence, co-segmentation, pose estimation, and 3D understanding~\cite{amir2021deep,goodwin2022zero,elbanani2024probing,goldblum2023battle,thrush2022winoground}, and similar correspondence emerges in diffusion features~\cite{tang2023emergent,luo2023diffusion}. However, correspondence metrics have been used primarily as post-hoc capability assessments, not as functional analysis tools to identify which training components give rise to semantic structure. We use SPair-71k~\cite{min2019spair} and NAVI~\cite{jampani2023navi} as zero-shot diagnostic probes across all ablation variants, and show that they predict dense downstream performance more reliably than classification proxies~(\cref{sec:scale}).

\section{Probing Semantic Emergence in DINO-Family SSL}
\label{sec:method}

\begin{figure}[t]
    \centering
    \includegraphics[width=\linewidth]{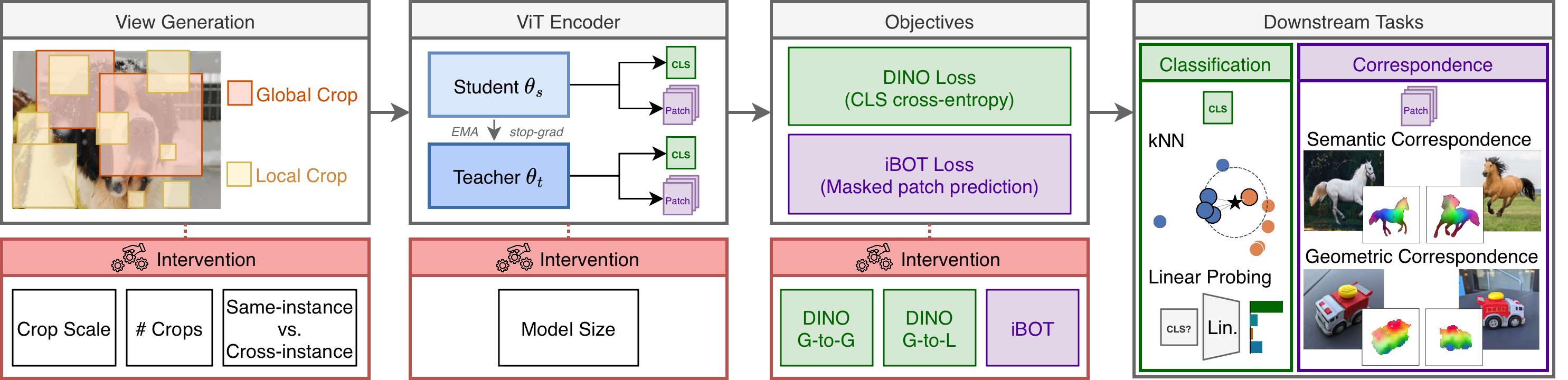}
%     \caption{%
%       \textbf{Experimental setup.}
%       We systematically intervene on three components of the DINO-style training pipeline to isolate the factors responsible for semantic emergence.
% (i) We study the \textit{view generation} to test the role of geometric displacement and instance identity in forming semantic correspondences.
% (ii) We study the \textit{training objective} to distinguish the contribution of global instance alignment (semantic anchoring) from patch-level masked prediction (semantic densification).
% (iii) We vary \textit{model capacity} to study how scale amplifies learned representations without altering the underlying learning mechanism.
% All configurations are evaluated along two complementary axes of representation quality: classification and correspondence.
% This intervention framework allows us to disentangle which components are necessary for semantic structure to emerge and which primarily refine or amplify it.
%     }
    \caption{%
      \textbf{Experimental setup.}
      We systematically intervene on three components of the DINO-style training pipeline to isolate the factors responsible for semantic emergence:
      (i) \textit{view generation} to test the role of geometric displacement and instance identity;
      (ii) \textit{training objective} to distinguish the contribution of global instance alignment (semantic anchoring) vs. patch-level masked prediction (semantic refinement);
      (iii) \textit{model capacity} to study how scale amplifies representations without altering the learning mechanism.
      All configurations are evaluated on classification and correspondence.
    }

    \label{fig:method}
\end{figure}

% ------------------------------------------------------------------
\subsection{DINO-Family Pretraining}\label{sec:method_dino_family}
% ------------------------------------------------------------------

\textbf{DINOv1}~\cite{caron2021emerging} uses a Vision Transformer~\cite{dosovitskiy2020vit} in a teacher--student framework~\cite{tarvainen2017mean} where student weights $\theta_s$ are updated by gradient descent and teacher weights $\theta_t$ are an EMA of $\theta_s$.
The encoder produces a global \texttt{[CLS]} token and spatial \emph{patch tokens}.
DINOv1 trains the student to match the teacher's
\texttt{[CLS]} token across differently augmented views of the same image, using
temperature-scaled softmax outputs and a cross-entropy loss with EMA centering to prevent collapse.
Two large global crops (scale $\in[s_{\min}, 1.0]$, default $s_{\min}{=}0.32$) and
eight small local crops (scale $\in[0.05, s_{\min}]$) are produced by the multi-crop strategy~\cite{caron2020swav}.
The teacher processes only global views; the student processes all views.
The DINO loss decomposes into a \emph{global-to-global} term and a \emph{local-to-global} term; isolating their semantic contributions is a key part of our analysis.

\textbf{DINOv2}~\cite{oquab2023dinov2} adds a patch-level self-distillation
objective from iBOT~\cite{zhou2021ibot}: a fraction of student input patches are
masked, and the student's features at masked positions are aligned with the
teacher's unmasked features via a per-patch cross-entropy loss with its own centering and softmax.
The full objective is $\mathcal{L}_\mathrm{DINOv2} = \mathcal{L}_\mathrm{DINO} + \mathcal{L}_\mathrm{iBOT}$,
which can be read as \emph{global-view \texttt{[CLS]} alignment + masked patch-level refinement}.

% ------------------------------------------------------------------
\subsection{Interventions}\label{sec:method_interventions}
% ------------------------------------------------------------------

\paragraph{Loss function.}
We compare the full DINOv2 objective (D+I), DINO only (D), and iBOT only
without any global contrastive signal (I).

\paragraph{View configuration.}
We vary the multi-crop schedule along three dimensions.
\emph{Global crop count}: from one (1G) to four (4G), varying the strength of the global contrastive signal.
\emph{Local crops}: removing all local crops (2G vs.\ 2G+8L) isolates whether local-to-global matching is necessary for semantic emergence.
\emph{Crop type}: photometric-only views apply distinct photometric augmentations to the \emph{same} spatial crop; cross-instance views pair crops from \emph{different images} within the same ImageNet synset.
These isolate which property of global-to-global contrast is strictly necessary.

\paragraph{Crop scale.}
We further sweep the global crop scale under two regimes using the DINO global-to-global loss: \textit{(i)} \emph{random-scale} sweep varying $s_{\min}$ in $[s_{\min},1]$, and \textit{(ii)} \emph{fixed-scale} sweep drawing both crops at the same scale $[s,s]$, which removes scale stochasticity entirely.

We do not intervene on photometric augmentations, as they are necessary for SSL pretraining at smaller scale~\cite{moutakanni2024augmentation}.

% ------------------------------------------------------------------
\subsection{Evaluation Protocol}\label{sec:method_eval}
% ------------------------------------------------------------------

\paragraph{Classification (\texttt{[CLS]} token).}
kNN top-1 ($k{=}10$) and linear probe top-1 on ImageNet-1k using frozen \texttt{[CLS]} features, following standard practice~\cite{chen2020simclr,caron2021emerging,oquab2023dinov2}.

\paragraph{Correspondence (patch tokens).}
Zero-shot nearest-neighbour matching on frozen patch features~\cite{amir2021deep,elbanani2024probing}.
\emph{SPair-71k}~\cite{min2019spair} measures cross-instance semantic (Sem.) correspondence via PCK@0.1.
\emph{NAVI}~\cite{jampani2023navi} measures within-instance geometric (Geom.) correspondence via R@0.05m.
We further validate both as proxies for dense downstream task performance via correlation analysis in~\cref{sec:scale}; full downstream task details are in~\cref{suppsec:downstream_task_details}.

\section{Experiments}
\label{sec:experiments}

\paragraph{Setup.}
All models use a ViT-L/16 backbone trained for 100 epochs on ImageNet-1k under the low-compute regime of DINOv2~\cite{oquab2023dinov2}.
Unless stated otherwise, view generation follows the default multi-crop recipe described in~\cref{sec:method_dino_family}: two global crops at scale $[0.32,1.0]$ and eight local crops at scale $[0.05,0.32]$, with photometric augmentations.
Classification is evaluated via frozen-feature kNN ($k=10$) and linear probing on ImageNet-1k.
Semantic quality is measured via zero-shot patch-feature correspondence on SPair-71k (PCK@0.1) and NAVI (R@0.05m), as defined in~\cref{sec:method_eval}.
% Throughout, we use ``collapse'' to mean performance at or below the iBOT-only baseline (SPair ${\leq}12$); ``degradation'' for losses ${>}10$ pp from the global-only baseline; and ``preservation'' for losses ${<}5$ pp.

\subsection{Model Size}
\label{sec:scale_arch}

\begin{wraptable}{r}{0.50\textwidth}
\vspace{-1em}
\centering
\caption{%
  \textbf{Effect of model size.}
  DINOv2 pretraining on ImageNet-1k. ViT-g/14 does not improve over ViT-L/14, suggesting overfitting.
  $\dagger$~300 epochs, all others 100.
}
\label{tab:model_scale}
\setlength{\tabcolsep}{5pt}
\small
\begin{tabular}{l  cc  cc}
\toprule
& \multicolumn{2}{c}{\textit{Class.}} & \multicolumn{2}{c}{\textit{Corr.}} \\
\cmidrule(lr){2-3}\cmidrule(lr){4-5}
Model & kNN & Lin. & Sem. & Geom. \\
      & Top-1 & Top-1 & PCK@.1 & R@.05m \\
\midrule
ViT-S/16 & 68.96 & 68.18 & 24.82 & 71.71 \\
ViT-B/16 & 74.79 & 75.15 & 31.24 & 72.78 \\
ViT-L/16 & 80.39 & 80.63 & 42.51 & 77.12 \\
ViT-L/14 & \textbf{80.74} & \textbf{80.43} & \textbf{47.33} & \textbf{77.15} \\
ViT-g/14$^\dagger$ & 74.59 & 75.73 & 47.21 & 76.38 \\
\bottomrule
\end{tabular}
\vspace{-1em}
\end{wraptable}

Before ablating individual components, we first establish how model capacity affects classification and correspondence, and select a backbone that balances performance with the compute cost of running many ablations.
We train DINOv2 across five architectures (100 epochs each, except ViT-g/14 at 300 epochs; details in \cref{suppsec:dino_training_model_size}) and report results in~\cref{tab:model_scale}.
Correspondence is more sensitive to scale than classification: semantic correspondence nearly doubles from ViT-S/16 to ViT-L/14 ($24.8{\to}47.3$) while kNN gains only 12 points.
ViT-g/14 does not improve over ViT-L/14 on correspondence despite more parameters and longer training, suggesting overfitting to ImageNet-1k at this data scale.
We fix \textbf{ViT-L/16} for all experiments: it gives the best correspondence among patch-16 models at a favorable compute-to-performance trade-off.

\textbf{On the scale of our experiments.} This 100-epoch ViT-L/16 regime is the standard low-compute setting of DINOv2~\cite{oquab2023dinov2}, consistent with SimDINO~\cite{wu2025simplifying} and LeJEPA~\cite{balestriero2025lejepa}. The over 20 ablation configurations we study would be computationally prohibitive at the full DINOv2 data scale.

\begin{table}[t]
\centering
\caption{%
  \textbf{Component ablation.}
  ViT-L/16, ImageNet-1k, 125k iterations.
  Losses: D\,=\,DINO, I\,=\,iBOT; G\,=\,global crops, L\,=\,local crops.
  \textbf{Bold}: best per column.
  Subscripts: $\Delta$ vs.\ full-pipeline reference.
}
\label{tab:main}
\setlength{\tabcolsep}{4pt}
\small
\begin{tabular}{@{}l cccc cccc@{}}
\toprule
& \multicolumn{4}{c}{\textit{DINOv2 (D+I)}}
  & \multicolumn{4}{c@{}}{\textit{DINOv1 (D)}} \\
\cmidrule(lr){2-5}\cmidrule(l){6-9}
Crop schedule
  & kNN   & Lin.  & Sem.   & Geom.
  & kNN   & Lin.  & Sem.   & Geom. \\
& Top-1 & Top-1 & PCK@.1 & R@.05m
  & Top-1 & Top-1 & PCK@.1 & R@.05m \\
\midrule
2G+8L\enspace(full pipeline)
  & \textbf{80.4} & \textbf{80.6} & \textbf{42.5} & \textbf{77.1}
  & \textbf{79.1} & \textbf{79.3} & 32.0 & \textbf{73.9} \\
2G\enspace(global-only)
  & $72.9_{{\color{red}\downarrow}7.5}$
  & $74.1_{{\color{red}\downarrow}6.5}$
  & $36.8_{{\color{red}\downarrow}5.7}$
  & $74.3_{{\color{red}\downarrow}2.8}$
  & $72.4_{{\color{red}\downarrow}6.7}$
  & $72.6_{{\color{red}\downarrow}6.7}$
  & $31.2_{{\color{red}\downarrow}0.8}$
  & $72.0_{{\color{red}\downarrow}1.9}$ \\
4G\enspace(four-global)
  & $76.3_{{\color{red}\downarrow}4.1}$
  & $77.6_{{\color{red}\downarrow}3.0}$
  & $40.7_{{\color{red}\downarrow}1.8}$
  & $75.5_{{\color{red}\downarrow}1.6}$
  & $74.6_{{\color{red}\downarrow}4.5}$
  & $75.7_{{\color{red}\downarrow}3.6}$
  & $\textbf{32.9}_{{\color{OliveGreen}\uparrow}0.9}$
  & $73.7_{{\color{red}\downarrow}0.2}$ \\
\bottomrule
\end{tabular}
\end{table}

% ------------------------------------------------------------------
\subsection{Intervention Ladder: Progressive Component Analysis}
\label{sec:ladder}
% ------------------------------------------------------------------

We systematically remove components from the DINOv2 recipe one at a time and measure effects on both classification and correspondence (\cref{tab:main}).
A central finding is that the two axes respond differently to each removal, revealing a dissociation between what drives classification quality and what gives rise to semantic representations.

\paragraph{iBOT refines but does not create semantic structure.}
Dropping the iBOT patch-prediction loss while retaining the DINO objective reduces DINOv2 to DINOv1.
Classification drops modestly (kNN $80.4{\to}79.1$, linear $80.6{\to}79.3$), and correspondence degrades (Sem.\ $42.5{\to}32.0$, Geom.\ $77.1{\to}73.9$), but crucially does not collapse.
DINOv1 patch features retain strong semantic structure, a result invisible to classification metrics alone.
The 10.5 pp semantic correspondence gap is non-trivial: iBOT is a \textbf{substantial refiner}.
We characterize it as a refiner rather than a creator because global-only DINOv1 (two global crops, no local crops) already establishes structured correspondence (Sem.\ 31.2), while iBOT trained without the DINO loss collapses entirely (Sem.\ 6.8; \cref{tab:necessity}), demonstrating it cannot bootstrap semantic structure independently.
iBOT's quantitative contribution is substantial within a working pipeline, and its relative role may increase at larger data scales.

\paragraph{Local crops improve classification but not correspondence.}
Removing the eight local crops from DINOv1 while retaining only two global views produces a \textbf{striking asymmetric effect}.
Classification drops substantially (kNN: $79.1{\to}72.4$, linear: $79.3{\to}72.6$), yet semantic and geometric correspondence shift only marginally (Sem.\ $32.0{\to}31.2$, Geom.\ $73.9{\to}72.0$).
If local-to-global matching were the operative mechanism for semantic emergence, removing it should collapse correspondence, yet it does not.
\textbf{Local crops are not the source of semantic structure.}
They contribute to classification-oriented quality by increasing view diversity, but \textbf{the semantic content of patch features is established by global-view contrast alone}.
We refer to DINOv1 on two global crops with no local crops as the \emph{minimal configuration}.

\paragraph{Sanity check: additional global crops partially recover classification.}
If local crops primarily supply view diversity rather than a qualitatively distinct signal, substituting additional global views should partially recover the classification gap.
DINOv1 with four global crops reaches kNN 74.6 and linear 75.7, recovering roughly half the gap to full DINOv1, with essentially unchanged correspondence (Sem.\ 32.9, Geom.\ 73.7).
Notably, four-global DINOv1 (Sem.\ 32.9) marginally exceeds full DINOv1 (Sem.\ 32.0), suggesting that in the DINO-only regime local crops mildly reduce correspondence relative to additional global crops; this does not alter the conclusion that view count primarily benefits classification.
\textbf{Additional views benefit classification, but are not the cause of semantic emergence.}

% ------------------------------------------------------------------
\subsection{Necessity Tests: What Makes Global-View Contrast Work?}
\label{sec:necessity}
% ------------------------------------------------------------------
Having established two global same-instance crops as a sufficient minimal configuration, we now isolate which property of that signal is operative.
We design four targeted experiments, each disabling one candidate mechanism; results are in~\cref{tab:necessity}.

% \textbf{A note on view-count confounds.}
% View count differs across conditions: the iBOT-only condition uses 2 global crops; single-global uses 1G+8L; photometric-only uses 1 crop; cross-instance uses 4 crops.
% However, view-count effects across the 2G, 2G+8L, and 4G configurations produce semantic correspondence differences of at most $\sim$2 pp — an order of magnitude smaller than the severe drops observed when global-view structure, geometric displacement, or instance identity are removed.
% This asymmetry strongly suggests the drops are driven by the intended ablated variable, not view-count differences.
% A fully controlled design (matched view count per condition) remains desirable future work.

\begin{table}[t]
\centering
\caption{%
  \textbf{Necessity tests.}
  Each condition removes one property from the minimal two-global-view DINOv1 configuration,
  which serves as the reference (top row).
  Subscripts show the drop ${\color{red}\downarrow}$ from the reference.
  Losses: D\,=\,DINO, I\,=\,iBOT.
}
\label{tab:necessity}
\setlength{\tabcolsep}{4pt}
\small
\begin{tabular}{@{}l l l cccc@{}}
\toprule
Condition & Loss & Views
  & kNN   & Lin.  & Sem.   & Geom. \\
& & & Top-1 & Top-1 & PCK@.1 & R@.05m \\
\midrule
DINOv1 global-only\enspace\textit{(reference)} & D & 2G
  & 72.4 & 72.6 & 31.2 & 72.0 \\
\midrule
iBOT only (no DINO) & I & 2G
  & $1.8_{{\color{red}\downarrow}70.6}$ & $9.4_{{\color{red}\downarrow}63.2}$
  & $6.8_{{\color{red}\downarrow}24.4}$ & $56.4_{{\color{red}\downarrow}15.6}$ \\
No global contrast & D & 1G+8L
  & $47.0_{{\color{red}\downarrow}25.4}$ & $57.0_{{\color{red}\downarrow}15.6}$
  & $10.6_{{\color{red}\downarrow}20.6}$ & $58.8_{{\color{red}\downarrow}13.2}$ \\
No spatial displacement & D & 4\,aug.$^{*}$
  & $22.4_{{\color{red}\downarrow}50.0}$ & $38.5_{{\color{red}\downarrow}34.1}$
  & $20.1_{{\color{red}\downarrow}11.1}$ & $68.2_{{\color{red}\downarrow}3.8}$ \\
Cross-instance & D & 4G$^{\dagger}$
  & $31.0_{{\color{red}\downarrow}41.4}$ & $38.3_{{\color{red}\downarrow}34.3}$
  & $7.3_{{\color{red}\downarrow}23.9}$ & $57.3_{{\color{red}\downarrow}14.7}$ \\
\bottomrule
\multicolumn{7}{@{}l@{}}{\scriptsize $^{*}$Four photometric augmentations of the same spatial crop.}\\
\multicolumn{7}{@{}l@{}}{\scriptsize $^{\dagger}$Two crops each from two different same-class images (4 crops total).}\\
\end{tabular}
\end{table}

\paragraph{iBOT alone cannot bootstrap semantic structure.}
Training with iBOT but \emph{without} the DINO objective leads to near-complete collapse: kNN 1.8, Sem.\ 6.8.
Patch features carry no meaningful semantic structure, ruling out the possibility that patch-level prediction alone can bootstrap semantic representations.

\textbf{The global contrastive objective is a strict prerequisite.}
Importantly, iBOT applies its own centering and softmax to patch-token predictions, so training dynamics are not destabilized by missing centering; the collapse is not a stability artifact.
Rather, patch-level self-distillation, even with intact centering, is insufficient to bootstrap global semantic structure; it requires the CLS-level contrastive objective as a foundation.

\paragraph{Global-to-global contrast is necessary, not merely beneficial.}
Training with a single global crop and 8 local crops reduces kNN to 47.0 and linear accuracy to 57.0; semantic correspondence collapses to 10.6 and geometric correspondence to 58.8.
Even with full photometric augmentation, the absence of a global-to-global contrastive signal eliminates semantic structure.
\textbf{Two-view global alignment is necessary, not merely beneficial.}

\paragraph{Geometric displacement is the operative signal.}
We replace geometrically displaced global crops with four photometric augmentations of the \emph{same} global crop.
Classification remains above chance (kNN 22.4, linear 38.5), reflecting residual photometric invariance, but semantic correspondence degrades to 20.1 and geometric correspondence to 68.2.
\textbf{The operative signal is geometric: the model must align representations across spatially displaced views of the same instance.}

\paragraph{Same-instance identity is necessary.}
We replace same-image global crops with crops from two different images within the same ImageNet synset (4 total crops, two per instance).
Classification remains above chance (kNN 31.0, linear 38.3), yet semantic correspondence collapses to 7.3 and geometric correspondence falls to 57.3.
The cross-instance result (Sem.\ 7.3) is comparably low to the iBOT-only result (Sem.\ 6.8); both indicate collapse.
\textbf{Class-level semantic similarity cannot substitute for within-instance geometric alignment.}

\paragraph{Summary.}
%The four necessity tests converge on a single conclusion: semantic emergence requires \textbf{alignment across geometrically displaced, same-image global views.}
%Removing any of the four tested properties individually causes collapse or severe degradation of correspondence, while classification remains non-trivially above chance in all cases.
Three necessity tests converge on a single conclusion: semantic emergence requires alignment across geometrically displaced, same-image, global views. Removing global-to-global contrast (no-global-contrast condition), geometric displacement (no-spatial-displacement condition), or same-instance identity (cross-instance condition) each causes severe degradation while classification remains non-trivially above chance. A fourth experiment (iBOT-only) confirms that the iBOT objective cannot substitute for this signal: even with intact centering, patch-level self-distillation alone fails to bootstrap semantic structure.

% Specifically, removing the DINO loss (Sem.\ 6.8), global-to-global contrast (10.6), or same-instance identity (7.3) produces collapse (SPair ${\leq}12$); removing geometric displacement (20.1) produces severe degradation.

% ------------------------------------------------------------------
\subsection{Crop Scale Analysis: Classification and Correspondence Have Opposing Optima}
\label{sec:scale}
% ------------------------------------------------------------------
The global views in our minimal configuration are generated by
\texttt{RandomResizedCrop} with scale drawn uniformly from $[s_\mathrm{min}, 1]$.
The experiments in~\cref{sec:ladder} hold this schedule fixed; here we ask how sensitive classification and correspondence are to crop scale, and whether they respond in the same direction.
We study two complementary settings:
\begin{itemize}
  \item \textbf{Random scale $[s_\mathrm{min}, 1]$}: vary $s_\mathrm{min} \in \{0.16, \ldots, 0.84\}$, controlling scale-range diversity.
  \item \textbf{Fixed scale $[s, s]$}: both crops drawn at the same fixed scale $s$, removing scale stochasticity entirely and isolating the effect of crop \emph{size}.
\end{itemize}

\begin{figure}[t]
    \centering
    \begin{subfigure}[t]{0.49\linewidth}
        \centering
        \includegraphics[width=\linewidth]{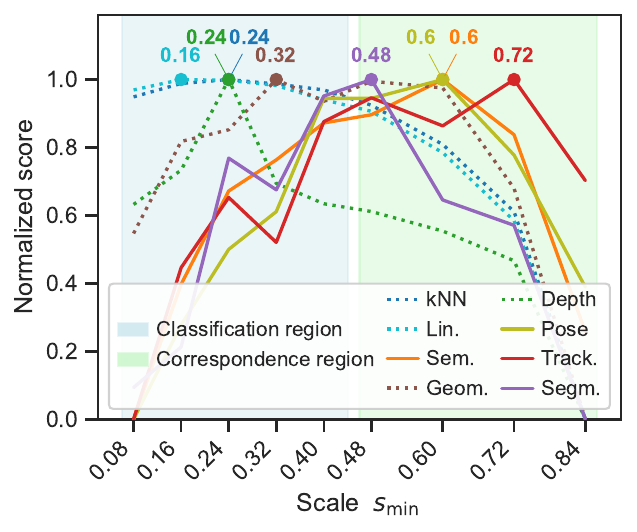}
        \caption{Random scale $[s_\mathrm{min},1]$}\label{fig:scale_curves_rand}
    \end{subfigure}
    \hfill
    \begin{subfigure}[t]{0.49\linewidth}
        \centering
        \includegraphics[width=\linewidth]{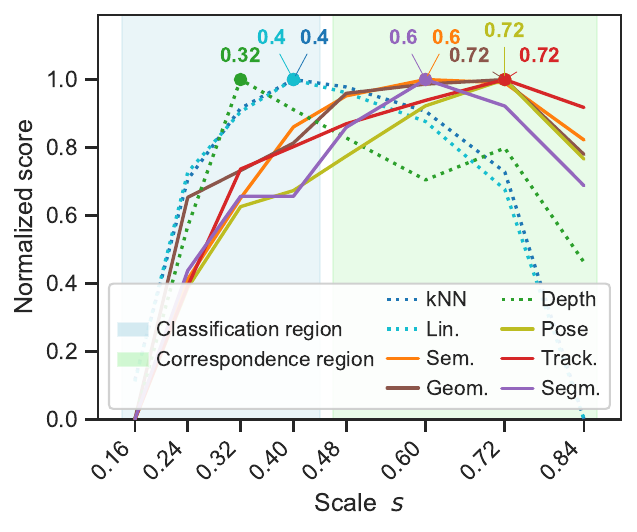}
        \caption{Fixed scale $[s,s]$}\label{fig:scale_curves_fixed}
    \end{subfigure}
    \caption{%
      \textbf{Classification and correspondence have opposing optima across crop scale.}
      All metrics normalized to $[0,1]$. Dotted lines (classification) peak at $s{=}0.40$;
      solid lines (correspondence) peak at $s{\geq}0.48$.
    }
    \label{fig:scale_curves}
    \vspace{-1em}
\end{figure}

All experiments use the DINOv1 loss on two global views with no local crops (the minimal configuration).
Results are shown in~\cref{tab:scale_cls_corr,tab:fixscale_cls_corr} and visualized in~\cref{fig:scale_curves_rand,fig:scale_curves_fixed}.

\begin{table}[t]
\centering
\caption{
\textbf{Crop scale ablation.}
DINOv1 loss, 2 global crops (no locals), ViT-L/16, 125k iterations. Sem. and Geom. refer to Semantic and Geometric correspondences.
}
\label{tab:scale_ablation_combined}
\setlength{\tabcolsep}{5pt}
\begin{subtable}[t]{0.48\linewidth}
\centering
\caption{\textbf{Random scale} $[s_{\min},1]$.}
\label{tab:scale_cls_corr}
\begin{tabular}{c  cc  cc}
\toprule
&
\multicolumn{2}{c}{\emph{Classification}} &
\multicolumn{2}{c}{\emph{Correspondence}} \\
\cmidrule(lr){2-3}\cmidrule(lr){4-5}
$s_{\min}$ &
kNN & Linear &
Sem. & Geom. \\
&
Top-1 & Top-1 &
PCK@.1 & R@.05m \\
\midrule
%0.08 & 70.56 & 72.00 & 25.62 & 72.55 \\
0.16 & 71.48 & \textbf{72.54} & 27.92 & 73.10 \\
0.24 & \textbf{71.75} & 72.50 & 29.51 & 73.17 \\
0.32 & 71.44 & 72.23 & 30.04 & \textbf{73.47} \\
0.40 & 71.04 & 71.49 & 30.67 & 73.34 \\
0.48 & 69.99 & 70.89 & 30.81 & 73.46 \\
0.60 & 67.30 & 68.75 & \textbf{31.41} & 73.42 \\
0.72 & 62.69 & 65.22 & 30.47 & 72.82 \\
0.84 & 48.45 & 54.92 & 27.11 & 71.44 \\
\bottomrule
\end{tabular}
\end{subtable}
\hfill
\begin{subtable}[t]{0.48\linewidth}
\centering
\caption{\textbf{Fixed scale} $[s,s]$.}
\label{tab:fixscale_cls_corr}
\begin{tabular}{c  cc  cc}
\toprule
&
\multicolumn{2}{c}{\emph{Classification}} &
\multicolumn{2}{c}{\emph{Correspondence}} \\
\cmidrule(lr){2-3}\cmidrule(lr){4-5}
$s$ &
kNN & Linear &
Sem. & Geom. \\
&
Top-1 & Top-1 &
PCK@.1 & R@.05m \\
\midrule
0.16 & 56.33 & 60.97 & 17.64 & 68.99 \\
0.24 & 65.43 & 68.05 & 23.22 & 71.89 \\
0.32 & 68.66 & 70.10 & 26.45 & 72.24 \\
0.40 & \textbf{69.97} & \textbf{71.19} & 29.29 & 72.60 \\
0.48 & 69.63 & 70.74 & 30.53 & 73.25 \\
0.60 & 68.53 & 69.77 & \textbf{31.18} & 73.37 \\
0.72 & 65.77 & 67.47 & 31.05 & \textbf{73.43} \\
0.84 & 54.54 & 59.69 & 28.77 & 72.45 \\
\bottomrule
\end{tabular}
\end{subtable}
\vspace{-1em}
\end{table}

\paragraph{Random scale: classification and correspondence peak differently.}
As $s_\mathrm{min}$ increases, the two global crops become more similar in size and spatial content.
kNN peaks at $s_\mathrm{min}=0.24$ and declines monotonically thereafter; by the correspondence optimum of $s_\mathrm{min}=0.60$ it has already dropped 4.5 points.
Semantic correspondence rises steadily to $s_\mathrm{min}=0.60$ before falling; geometric correspondence peaks earlier at $s_\mathrm{min}=0.32$.

\paragraph{Fixed scale: the dissociation holds without stochasticity.}
The split is cleaner in this setting.
kNN peaks at $s{=}0.40$ and declines monotonically as crops grow more overlapping.
Semantic correspondence peaks later at $s{=}0.60$ and geometric correspondence later still at $s{=}0.72$, well past the classification optimum.
This reflects a genuine disentanglement between what each objective requires from crop geometry.

\paragraph{Correspondence tracks downstream tasks better.}

The normalized performance curves in~\cref{fig:scale_curves} overlay all eight metrics for both scale regimes.
In the fixed-scale panel (\cref{fig:scale_curves_fixed}), the split is clean: kNN and Lin.\ peak at $s=0.40$, while correspondence and downstream tasks (depth RMSE, pose $\pi/6$, tracking TAPVid avg-$\delta$, and segmentation ADE20k mIoU) peak at $s \geq 0.60$.
The random-scale panel is noisier; we attribute this to the confound between scale diversity and mean crop size.
The fixed-scale experiment provides the cleaner, more diagnostic signal.
\begin{figure}[t]
    \centering
    \begin{subfigure}[t]{0.32\linewidth}
        \centering
        \includegraphics[width=\linewidth]{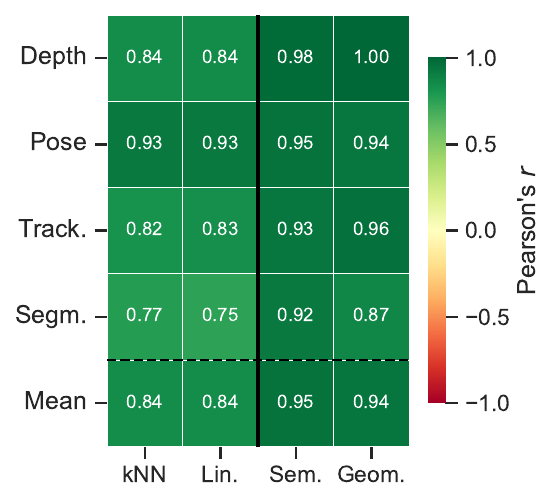}
        \caption{Component interventions}\label{fig:proxy_heatmap_cat}
    \end{subfigure}
    \hfill
    \begin{subfigure}[t]{0.32\linewidth}
        \centering
        \includegraphics[width=\linewidth]{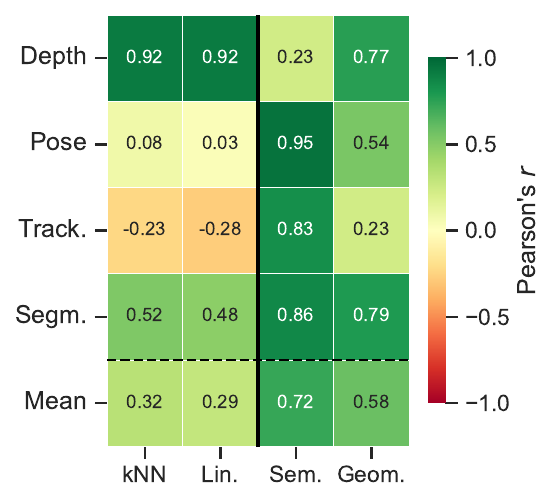}
        \caption{Random scale $[s_{\min},1]$}\label{fig:proxy_heatmap_rand}
    \end{subfigure}
    \hfill
    \begin{subfigure}[t]{0.32\linewidth}
        \centering
        \includegraphics[width=\linewidth]{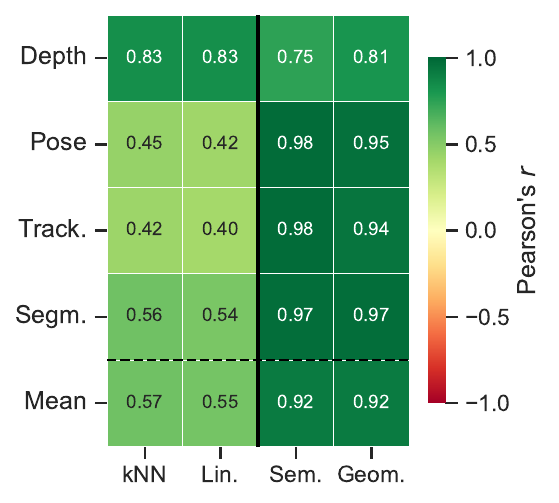}
        \caption{Fixed scale $[s,s]$}\label{fig:proxy_heatmap_fixed}
    \end{subfigure}
    \caption{%
      \textbf{Correspondence metrics better predict dense downstream performance
      than classification accuracy.}
      Pearson's $r$ between each proxy (kNN, Lin, Sem., Geom.) and four dense
      tasks (Depth, Pose, Tracking, Segmentation) across component interventions,
      random-scale sweep, and fixed-scale sweep.
      See \cref{fig:bootstrap-correlation-ci} in the supplementary for 95\% bootstrap CIs.
    }
    \label{fig:proxy_heatmap}
    \vspace{-1em}
\end{figure}

Bootstrap 95\% CIs (10{,}000 resamples over variants) confirm this trend holds across all three experimental settings; see \cref{fig:bootstrap-correlation-ci} in the supplementary.

Taken together, the scale experiments deliver two conclusions.
First, the \textbf{operative inductive bias for semantic correspondence is the geometric alignment of spatially overlapping views}, maximized at larger crop scales.
Second, this property is a better predictor of dense downstream utility than classification accuracy.

\section{Discussion and Limitations}
\label{sec:discussion}
Same-instance geometric view alignment is the operative inductive bias for semantic emergence; iBOT and local crops are engineering refinements rather than its source.
The DINO loss alone recovers most of the correspondence quality of the full DINOv2 recipe (75\% of semantic PCK and 93\% of geometric R@.05).

Several limitations bound the scope of these conclusions.
How the relative contribution of each component scales with data size and training duration remains open; the relative gains we identify may shift at larger scale and data diversity, even if the primacy of view-alignment as the operative inductive bias holds.
Our analysis is also specific to DINO-style teacher--student SSL; whether the same inductive bias governs semantic emergence in other SSL families remains an open question.
We do not analyse DINOv3~\cite{simeoni2025dinov3}, which scales beyond our fixed compute budget.
We also do not ablate batch size or self-distillation dynamics (EMA, centering), as these govern training stability and cannot be removed without confounding the training regime itself; we leave their interaction with semantic emergence to future work.

\section{Conclusion}
\label{sec:conclusion}

We have presented a systematic analysis of semantic emergence in DINO-style SSL, showing that same-instance global-view alignment is the semantic anchor of DINO-style learning, sufficient to produce strong semantic patch features and necessary for them to emerge. iBOT refines this structure (10.5 PCK points) but does not create it; local crops contribute view diversity that benefits classification but not correspondence; and necessity tests confirm that geometric displacement and same-instance identity are both strictly required.
As a methodological contribution, we show that semantic and geometric correspondence metrics (evaluated on SPair-71k and NAVI) track downstream performance more faithfully than classification accuracy, and advocate for their routine inclusion in SSL evaluation.
Together, these findings clarify what DINO-style training actually learns, offer a simplified recipe that preserves correspondence quality, and provide a functional analysis for evaluating future SSL methods.

% ---- Bibliography ----
\bibliographystyle{plainnat}
\bibliography{main}
%%%%%%%%%%%%%%%%%%%%%%%%%%%%%%%%%%%%%%%%%%%%%%%%%%%%%%%%%%%%

\appendix

\newpage
\maketitlesupplementary
\appendix

\renewcommand{\thefigure}{A\arabic{figure}}
\renewcommand{\thetable}{A\arabic{table}}
\renewcommand{\theequation}{A\arabic{equation}}
\setcounter{figure}{0}
\setcounter{table}{0}
\setcounter{equation}{0}

\newcommand{\additem}[2]{%
\item[\textbf{(\ref{#1})}] 
    \textbf{#2} \dotfill\makebox{\textbf{\pageref{#1}}
    }
}
\newcommand{\myindent}{.5em}
\newcommand{\addsubitem}[2]{%
\vspace{.2em}
    \textbf{(\ref{#1})}
        \hspace{\myindent} #2 \\    
}

\setlist[itemize]{noitemsep,leftmargin=*,topsep=0em}
\setlist[enumerate]{noitemsep,leftmargin=*,topsep=0em}

\noindent This supplementary provides additional implementation and training details
(\cref{suppsec:dino_training}), precise definitions of all downstream evaluation
tasks (\cref{suppsec:downstream_task_details}), qualitative feature visualizations
(\cref{suppsec:qualitative_analysis}), and bootstrap confidence intervals for the
correlation analysis (\cref{suppsec:bootstrap_ci}).
\vspace{0.3in}

\begin{enumerate}[label={({\arabic*})}, topsep=1em, itemsep=.2em]
    \additem{suppsec:dino_training}{DINOv2 training}\\[0.4em]\hfill

    \additem{suppsec:downstream_task_details}{Downstream task details}\\[0.4em]\hfill

    \additem{suppsec:qualitative_analysis}{Qualitative analysis}\\[0.4em]\hfill

    \additem{suppsec:bootstrap_ci}{Bootstrap confidence intervals}\\[0.4em]\hfill

\end{enumerate}

\newpage
\section{DINOv2 training}
\label{suppsec:dino_training}
\subsection{Implementation details}
\label{suppsec:implementation_details}
\paragraph{Architecture.}
All models use a Vision Transformer~\cite{dosovitskiy2020vit} trained with the
DINOv2 codebase~\cite{oquab2023dinov2} in the low-compute regime.
Our primary ablation backbone is ViT-L/16 (patch size 16, global crop size 224).
For the model size study, we additionally train ViT-S/16, ViT-B/16, ViT-L/14,
and ViT-g/14; the latter uses patch size 14 and a local crop size of 98 pixels
instead of 96.

\paragraph{Dataset.}
All models are trained exclusively on ImageNet-1k~\cite{imagenet15russakovsky},
in contrast to the full DINOv2 recipe, which uses a curated dataset of 142M
images~\cite{oquab2023dinov2}.
This is a deliberate choice: restricting training to a single, well-controlled
dataset allows us to isolate the effects of architectural and augmentation
choices without confounds from data curation or scale.

\paragraph{Training.}
All experiments use a total batch size of 1024 and a cosine learning rate
schedule with warmup, following the DINOv2 defaults except where noted.
The 2-global-crop ablations (DINOv1, DINOv2, and all necessity tests) use a
base learning rate of $4\times10^{-3}$; the 4-global-crop configurations and
all crop-scale sweep experiments use $2\times10^{-3}$.
All component ablations and the model size study train for 100 epochs.
Since no standard 100-epoch schedule exists for ViT-g/14 on ImageNet-1k, we
follow the large-scale DINOv2 recipe for this model: 300 epochs with a base
learning rate of $2\times10^{-4}$.
ViT-g/14 is excluded from all component ablations; results appear only in the
model size comparison (\cref{tab:model_scale}).

\subsection{Model size}
\label{suppsec:dino_training_model_size}
% \aj{Here explain why DINO-g has to be trained for 300 epochs }
% \begin{figure}[t]
%   \centering
%   \includegraphics[width=0.6\linewidth]{figures/knn_vs_spair_supp.jpg}
%   \caption{%
%     \textbf{kNN Top-1 vs.\ SPair-71k PCK@0.1 per checkpoint} for three
%     DINOv2 variants trained on ImageNet-1k (each point = one checkpoint,
%     curves progress from 12k to 124k/374k iterations).
%     ViT-G/14's kNN saturates at the same level as ViT-L models (overfitting
%     ImageNet-1k), and despite $3\times$ more training never surpasses ViT-L/14
%     on SPair; ViT-L/16 is selected over ViT-L/14 as the best
%     compute-to-performance trade-off.
%   }
%   \label{fig:knn_vs_spair}
% \end{figure}

\begin{figure}[t]
  \centering
  \begin{subfigure}[b]{0.48\linewidth}
    \includegraphics[width=\linewidth]{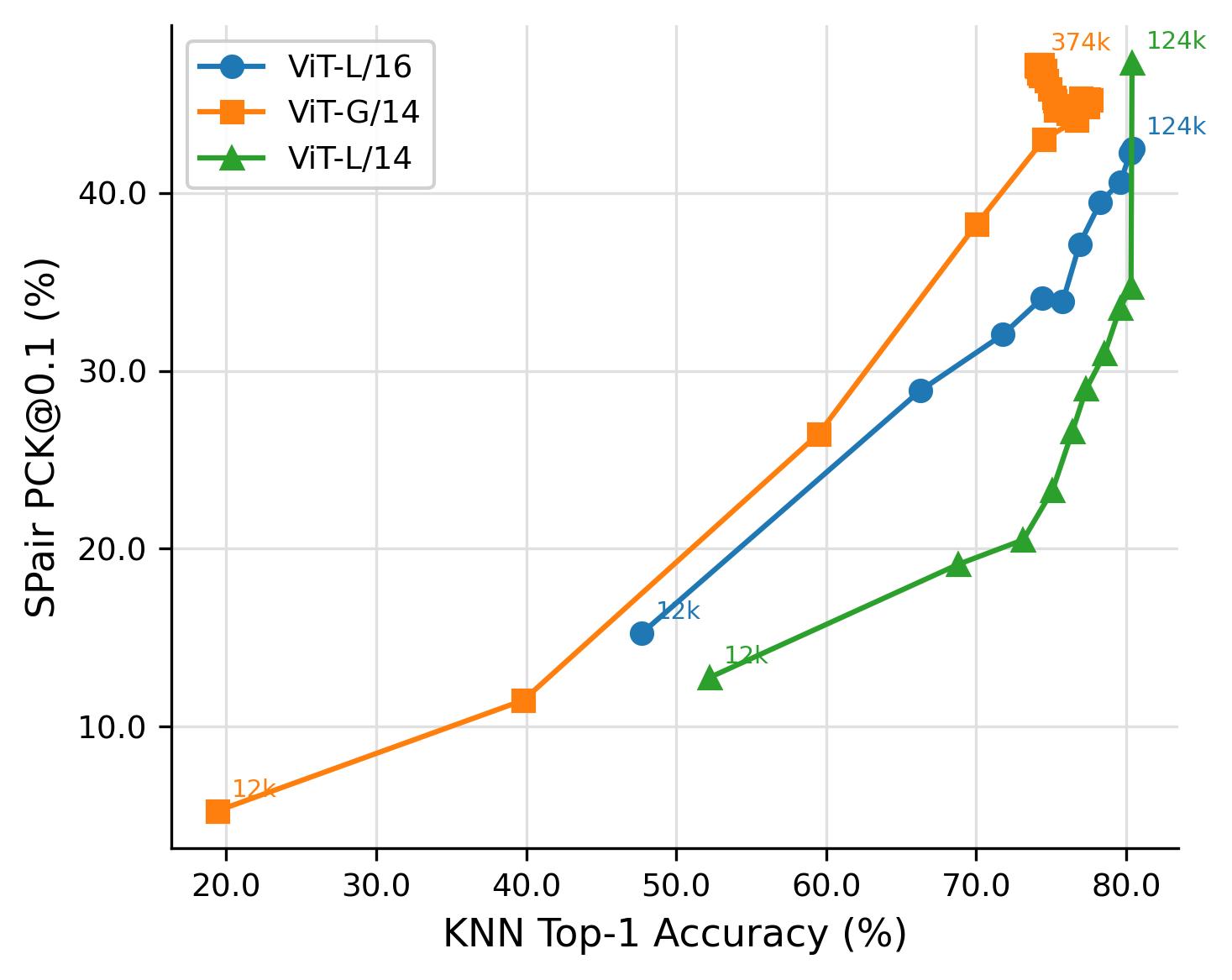}
    \caption{Classification vs.\ Semantic Correspondence}
    \label{fig:knn_vs_spair}
  \end{subfigure}
  \hfill
  \begin{subfigure}[b]{0.48\linewidth}
    \includegraphics[width=\linewidth]{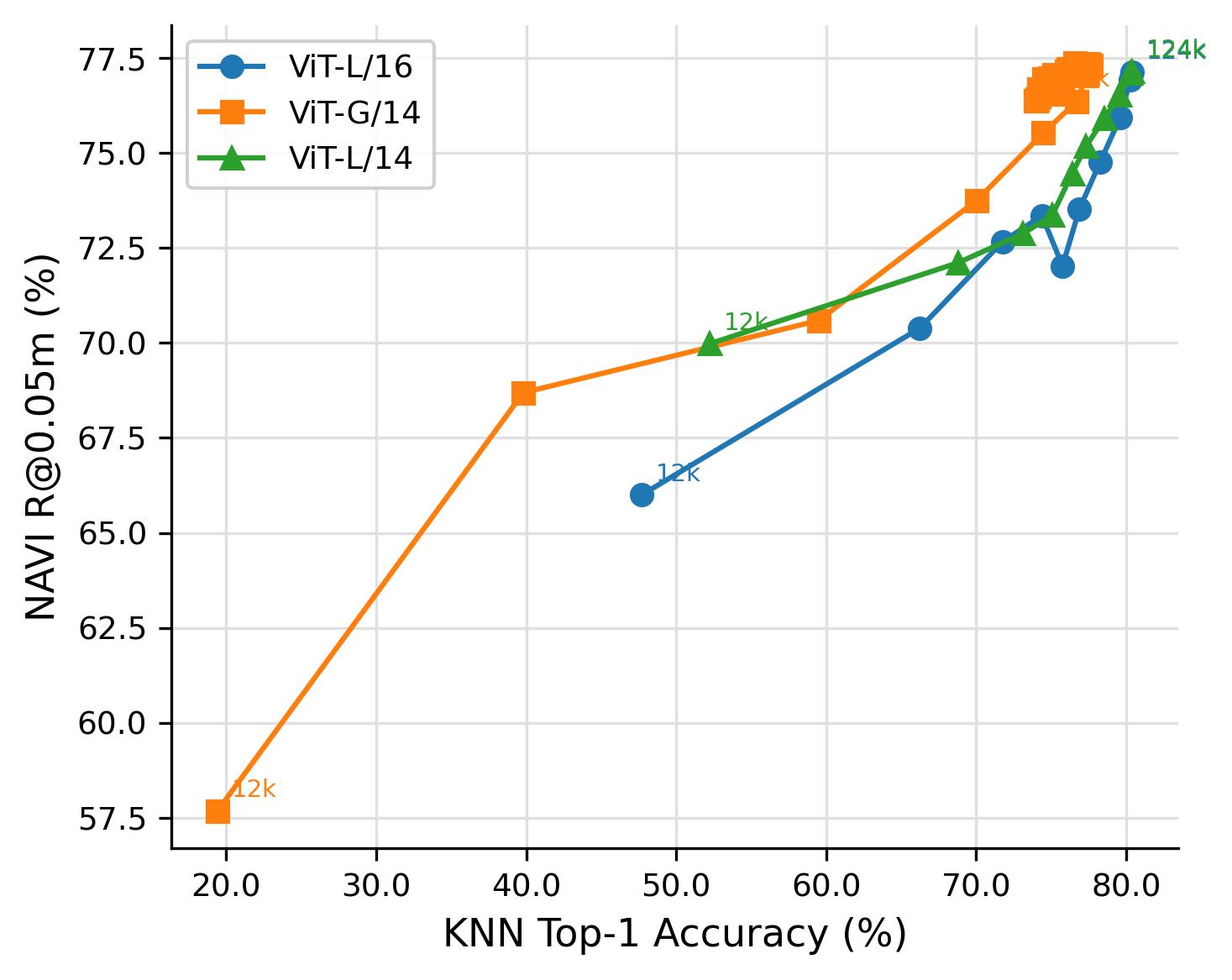}
    \caption{Classification vs.\ Geometric Correspondence}
    \label{fig:knn_vs_navi}
  \end{subfigure}
  \caption{%
    \textbf{kNN Classification accuracy vs.\ correspondence metrics throughout training}
    for three DINOv2 variants trained on ImageNet-1k (each point = one
    checkpoint, curves progress from 12k (epoch 10) to 124k/374k (epoch 100/300) iterations).
    ViT-G/14's classification performance saturates at the same level as the ViT-L models, indicating
    overfitting to ImageNet-1k; despite $3\times$ more iterations it never surpasses ViT-L/14 on semantic correspondence and yields no gain on geometric correspondence beyond 124k iterations.
    ViT-L/16 matches or closely approaches ViT-L/14 on both metrics at equal training budget, making it the best compute-to-performance choice (patch
    size 16 yields fewer tokens at a modest $\sim$5\% cost for semantic correspondence).
  }
  \label{fig:knn_vs_correspondence}
\end{figure}
% To select the model architecture for our experiments, we train three
% DINOv2 variants (ViT-L/16, ViT-L/14, and ViT-g/14) on ImageNet-1k and
% evaluate each checkpoint on ImageNet-1k kNN classification (Top-1, $k{=}10$)
% and SPair-71k semantic correspondence (PCK@0.1); see \cref{fig:knn_vs_correspondence}.
% % All three models exhibit a strong positive correlation between the two metrics
% % throughout training, confirming that a single self-supervised objective
% % simultaneously improves both recognition and geometric correspondence.
% Despite its substantially larger capacity, ViT-g/14's kNN accuracy saturates
% at the same level as the ViT-L models, suggesting that the model overfits
% to the ImageNet-1k classification task.
% Moreover, even after $3\times$ more training iterations (374k vs.\ 124k),
% ViT-g/14 never surpasses ViT-L/14 on SPair ($\sim$47.2\% vs.\ 47.3\%),
% confirming that the additional capacity does not translate into better
% geometric correspondence at this data scale.

To select the model architecture for our experiments, we train three
DINOv2 variants (ViT-L/16, ViT-L/14, and ViT-g/14) on ImageNet-1k and
evaluate each checkpoint on ImageNet-1k kNN classification (Top-1, $k$=10),
SPair-71k semantic correspondence (PCK@0.1), and NAVI geometric correspondence
(R@0.05m); see \cref{fig:knn_vs_correspondence}.
% All three models exhibit a strong positive correlation between kNN and both
% correspondence metrics throughout training.
Despite its substantially larger capacity, ViT-g/14's kNN accuracy saturates
at the same level as the ViT-L models, indicating that the model overfits
the ImageNet-1k classification task.
Strikingly, even after $3\times$ more training iterations (374k vs.\ 124k),
ViT-g/14 never surpasses ViT-L/14 on SPair ($\sim$47.2\% vs.\ 47.3\%),
and its additional training beyond 124k iterations yields no improvement on
NAVI either, confirming that additional compute does not translate into better
correspondence at this data scale
On NAVI, all three models converge to $\sim$77\% R@0.05m by 124k
iterations; ViT-g/14 starts lower (57.7\% at 12k iterations) but catches
up by $\sim$100k, after which further training yields no additional gain.
Meanwhile, the coarser patch size of ViT-L/16 reduces the number of tokens per
image, yielding better training throughput at only a modest
$\sim$5\% reduction in SPair performance (42\% vs.\ 47\%).
Based on these results, we adopt \textbf{ViT-L/16} for all subsequent experiments.

% \subsection{Training/validation curves}
% \label{suppsec:training_curves}

\section{Downstream task details}
\label{suppsec:downstream_task_details}
\subsection{Classification}\label{suppsec:cls}
We evaluate classification quality from frozen \texttt{[CLS]} token features
on the ImageNet-1k~\cite{imagenet15russakovsky} validation set.

\paragraph{kNN.}
Each validation image is classified via $k$-nearest-neighbour lookup ($k{=}10$)
over the full ImageNet-1k training set using cosine similarity on frozen
\texttt{[CLS]} features, with no fine-tuning.
We report Top-1 accuracy.

\paragraph{Linear.}
A single linear layer is trained on top of the frozen \texttt{[CLS]} token for
100 epochs with a cosine learning rate schedule, following standard
protocol~\cite{caron2021emerging,oquab2023dinov2}.
We report Top-1 accuracy on the validation set.

\subsection{Correspondence}\label{suppsec:corresp}
We evaluate the spatial structure of frozen patch features via zero-shot
nearest-neighbour matching, following~\cite{elbanani2024probing}.
For each query point, its patch feature is extracted and matched to the target
feature map by maximum cosine similarity.
No training or fine-tuning is performed at any stage.
 
\paragraph{Semantic correspondence (Sem.).}
SPair-71k~\cite{min2019spair} provides annotated semantic keypoint pairs across
18 object categories, covering variation in viewpoint, scale, truncation, and
occlusion.
We follow the zero-shot evaluation protocol of~\cite{elbanani2024probing}: the
source keypoint is assigned to the patch with highest cosine similarity to its
query feature in the target image.
We report PCK@0.1: the fraction of predicted keypoints falling within
$0.1{\times}\max(H,W)$ of the ground-truth location.
 
\paragraph{Geometric correspondence (Geom.).}
NAVI~\cite{jampani2023navi} provides multi-view captures of 36 object instances
with ground-truth 3D geometry and calibrated cameras.
Correspondences are established by cosine-similarity matching; ground-truth correspondences are derived from the known camera geometry
and 3D reprojection, following~\cite{elbanani2024probing}.
We report R@0.05\,m: the fraction of correspondences with a 3D reprojection
error below 5\,cm.

\subsection{Geometry}\label{suppsec:geometry}
We assess whether patch features support geometry estimation, following the
linear-probing framework of~\cite{elbanani2024probing}.

\paragraph{Depth.}
We assess monocular depth estimation following the linear-probing framework
of~\cite{elbanani2024probing}.
A DPT-style decoder is trained on top of frozen patch features extracted from
the last three ViT transformer blocks on NYU Depth v2~\cite{silberman2012indoor}.
The decoder fuses the three feature maps via feature-fusion blocks with bilinear
upsampling; all backbone weights are frozen.
We report RMSE (lower is better).

\paragraph{Pose estimation.}
Three linear classification heads are trained on top of frozen backbone features to
predict 3D object pose on ImageNet3D~\cite{ma2024imagenet3d}.
Each head independently predicts discretized azimuth, elevation, and in-plane
rotation; all backbone weights are frozen.
We report Acc@$\pi/6$: the fraction of test images whose predicted pose is within
$\pi/6$ radians ($30^\circ$) geodesic distance on SO(3) from the ground truth.

\subsection{Temporal}\label{suppsec:temporal}
We evaluate the ability of frozen patch features to track keypoints
across video frames.
Following~\cite{aydemir2024can}, we apply a zero-shot cost-volume approach on
TAP-Vid DAVIS~\cite{doersch2022tapvid}: for each query point, its patch feature
is extracted at the query frame and matched to subsequent frames by minimum
feature distance via low-capacity convolutional branches (5.5K parameters).
We report $\delta^\mathrm{vis}_\mathrm{avg}$: the average fraction of visible
tracked points within each of five pixel-distance thresholds
$\{1,2,4,8,16\}$ of the ground-truth location, evaluated only on visible points.
%\aj{Why is \textsc{b}\oldstylenums{3} better than \textsc{a}\oldstylenums{1} in point tracking task, i think this would be worth be explained here }

\subsection{Semantic Segmentation}\label{suppsec:segmentation}
We evaluate semantic segmentation on ADE20K~\cite{zhou2017scene}
(150 semantic categories, 20,210 training images) using a linear probe on top
of frozen patch features from the final ViT layer, following the protocol
of~\cite{pariza2025near}.
A single linear head is trained on spatial patch features; predictions are
bilinearly upsampled to the input resolution.
All backbone weights are frozen.
We report mIoU on the ADE20K validation set.

\begin{table*}[t]
\centering
\caption{%
  \textbf{Complete intervention results: all experiments and all metrics.}
  All models: ViT-L/16, ImageNet-1k, 125k iterations.
  Losses: D\,=\,DINO, I\,=\,iBOT; G\,=\,global crops, L\,=\,local crops.
  \textbf{Bold}: best per column.
  All metrics higher-is-better except RMSE$\downarrow$ (cm).
}
\label{tab:full_paper}
\setlength{\tabcolsep}{3pt}
\small
\begin{tabular}{@{}l l l  cc  cc  cc  c  c@{}}
\toprule
& & &
\multicolumn{2}{c}{\textit{Classification}} &
\multicolumn{2}{c}{\textit{Correspondence}} &
\multicolumn{2}{c}{\textit{Geometry}} &
\textit{Temporal} &
\textit{Segm.} \\
\cmidrule(lr){4-5}\cmidrule(lr){6-7}\cmidrule(lr){8-9}\cmidrule(lr){10-10}\cmidrule(lr){11-11}
Condition & Loss & Views &
kNN & Lin. &
Sem. & Geom. &
Depth & Pose &
Track. &
Sem. \\
& & &
Top-1 & Top-1 &
PCK@.1 & R@.05m &
RMSE & $\pi/6$ &
avg-$\delta$ &
mIoU \\
\midrule
\multicolumn{11}{@{}l@{}}{\textit{Full pipelines}} \\[2pt]
DINOv2
  & D+I & 2G+8L
  & \textbf{80.4} & \textbf{80.6}
  & \textbf{42.5} & \textbf{77.1}
  & \textbf{35.5} & \textbf{47.6}
  & 34.2
  & \textbf{28.7} \\
DINOv1
  & D & 2G+8L
  & 79.1 & 79.3
  & 32.0 & 73.9
  & 39.6 & 43.7
  & 34.2
  & 16.4 \\
\midrule
\multicolumn{11}{@{}l@{}}{\textit{Intervention ladder: progressive component analysis}} \\[2pt]
DINOv2, $\varnothing$L
  & D+I & 2G
  & 72.9 & 74.1
  & 36.8 & 74.3
  & 40.3 & 45.8
  & 33.0
  & 27.1 \\
DINOv1, $\varnothing$L
  & D & 2G
  & 72.4 & 72.6
  & 31.2 & 72.0
  & 42.5 & 42.8
  & 32.2
  & 18.2 \\
DINOv2, 4G
  & D+I & 4G
  & 76.3 & 77.6
  & 40.7 & 75.5
  & 37.2 & 45.3
  & \textbf{35.4}
  & 22.9 \\
DINOv1, 4G
  & D & 4G
  & 74.6 & 75.7
  & 32.9 & 73.7
  & 40.2 & 42.8
  & 34.6
  & 13.5 \\
\midrule
\multicolumn{11}{@{}l@{}}{\textit{Necessity tests: isolating the operative signal}} \\[2pt]
iBOT only (no DINO)
  & I & 2G
  & 1.8 & 9.4
  & 6.8 & 56.4
  & 61.0 & 31.1
  & 15.9
  & 6.8 \\
No global contrast
  & D & 1G+8L
  & 47.0 & 57.0
  & 10.6 & 58.8
  & 58.8 & 36.7
  & 21.1
  & 6.8 \\
No spatial displacement
  & D & 4\,aug.$^{*}$
  & 22.4 & 38.5
  & 20.1 & 68.2
  & 46.1 & 38.6
  & 28.1
  & 10.4 \\
Cross-instance
  & D & 4G$^{\dagger}$
  & 31.0 & 38.3
  & 7.3 & 57.3
  & 58.0 & 37.3
  & 10.1
  & 5.5 \\
\bottomrule
\multicolumn{11}{@{}l@{}}{\scriptsize $^{*}$Four photometric augmentations of the same spatial crop.} \\
\multicolumn{11}{@{}l@{}}{\scriptsize $^{\dagger}$Two crops each from two different same-class images (4 crops total).} \\
\end{tabular}
\end{table*}

% Paper table — 8-metric subset of tab_scale_full.tex.
% Dropped: ScanNet, Depth SI-d1, SNorm d1, SNorm RMSE, MC Det.
% Kept: KNN, Lin, SPair, NAVI, Depth RMSE, Pose, TAPVid, ADE20K.

\begin{table*}[t]
\centering
\caption{%
  \textbf{Crop scale ablation (random scale $[s_{\min}, 1]$): complete metrics.}
  All experiments use DINOv1 loss with 2 global crops (no locals), ViT-L/16, 125k iters.
  At each training step the minimum crop scale is sampled from $[s_{\min}, 1]$.
  All metrics are higher-is-better except RMSE$\downarrow$ (lower-is-better), reported in cm.
Pose accuracy is reported as the fraction of samples below a pose error threshold of $\pi/6$.
}
\label{tab:scale_paper}
\setlength{\tabcolsep}{3pt}
\scriptsize
\begin{tabular}{c  cc  cc  cc  c  c}
\toprule
&
\multicolumn{2}{c}{\emph{Classification}} &
\multicolumn{2}{c}{\emph{Correspondence}} &
\multicolumn{2}{c}{\emph{Geometry}} &
\emph{Temporal} &
\emph{Segment.} \\
\cmidrule(lr){2-3}\cmidrule(lr){4-5}\cmidrule(lr){6-7}\cmidrule(lr){8-8}\cmidrule(lr){9-9}
$s_{\min}$ &
kNN & Lin &
Sem. & Geom. &
Depth & Pose &
Track. &
Sem. \\
 &
Top-1 & Top-1 &
R@.1 & R@.05m &
RMSE$\downarrow$ & $\pi$/6 &
avg-$\delta$ &
mIoU \\
\midrule
% 0.08 & 70.56 & 72.00 & 25.62 & 72.55 & 0.4302 & 40.2 & 31.43 & 12.11 \\
0.16 & \second{71.48} & \textbf{72.54} & 27.92 & 73.10 & 42.58 & 40.7 & 32.51 & 12.43 \\
0.24 & \textbf{71.75} & \second{72.50} & 29.51 & 73.17 & \textbf{41.42} & 41.1 & 33.01 & 13.92 \\
0.32 & 71.44 & 72.23 & 30.04 & \textbf{73.47} & 42.75 & 41.3 & 32.69 & 13.67 \\
0.40 & 71.04 & 71.49 & 30.67 & 73.34 & 43.01 & \second{41.9} & 33.55 & \second{14.41} \\
0.48 & 69.99 & 70.89 & 30.81 & \second{73.46} & 43.11 & \second{41.9} & \second{33.72} & \textbf{14.54} \\
0.60 & 67.30 & 68.75 & \textbf{31.41} & 73.42 & 43.36 & \textbf{42.0} & 33.52 & 13.59 \\
0.72 & 62.69 & 65.22 & \second{30.47} & 72.82 & 43.74 & 41.6 & \textbf{33.85} & 13.39 \\
0.84 & 48.45 & 54.92 & 27.11 & 71.44 & 45.77 & 40.9 & 33.13 & 11.86 \\
\bottomrule
\end{tabular}
\end{table*}

% Paper table — 8-metric subset of tab_fixscale_full.tex.
% Dropped: ScanNet, Depth SI-d1, SNorm d1, SNorm RMSE, MC Det.
% Kept: KNN, Lin, SPair, NAVI, Depth RMSE, Pose, TAPVid, ADE20K.

\begin{table*}[t]
\centering
\caption{%
  \textbf{Crop scale ablation (fixed scale $[s, s]$): complete metrics.}
  All experiments use DINOv1 loss with 2 global crops (no locals), ViT-L/16, 125k iters.
  Both views are cropped at the same fixed scale $s$ (no scale randomness).
  All metrics are higher-is-better except RMSE$\downarrow$ (lower-is-better), reported in cm.
Pose accuracy is reported as the fraction of samples below a pose error threshold of $\pi/6$.
}
\label{tab:fixscale_paper}
\setlength{\tabcolsep}{3pt}
\scriptsize
\begin{tabular}{c  cc  cc  cc  c  c}
\toprule
&
\multicolumn{2}{c}{\emph{Classification}} &
\multicolumn{2}{c}{\emph{Correspondence}} &
\multicolumn{2}{c}{\emph{Geometry}} &
\emph{Temporal} &
\emph{Segment.} \\
\cmidrule(lr){2-3}\cmidrule(lr){4-5}\cmidrule(lr){6-7}\cmidrule(lr){8-8}\cmidrule(lr){9-9}
$s$ &
kNN & Lin &
Sem. & Geom. &
Depth & Pose &
Track. &
Sem. \\
 &
Top-1 & Top-1 &
R@.1 & R@.05m &
RMSE$\downarrow$ & $\pi$/6 &
avg-$\delta$ &
mIoU \\
\midrule
0.16 & 56.33 & 60.97 & 17.64 & 68.99 & 47.08 & 37.17 & 27.06 &  8.20 \\
0.24 & 65.43 & 68.05 & 23.22 & 71.89 & 43.87 & 38.97 & 29.68 & 11.00 \\
0.32 & 68.66 & 70.10 & 26.45 & 72.24 & \textbf{41.43} & 40.08 & 31.98 & 12.40 \\
0.40 & \textbf{69.97} & \textbf{71.19} & 29.29 & 72.60 & \second{41.93} & 40.30 & 32.42 & 12.40 \\
0.48 & \second{69.63} & \second{70.74} & 30.53 & 73.25 & 42.40 & 40.77 & 32.87 & 13.70 \\
0.60 & 68.53 & 69.77 & \textbf{31.18} & \second{73.37} & 43.10 & \second{41.46} & \second{33.33} & \textbf{14.60} \\
0.72 & 65.77 & 67.47 & \second{31.05} & \textbf{73.43} & 42.57 & \textbf{41.82} & \textbf{33.74} & \second{14.10} \\
0.84 & 54.54 & 59.69 & 28.77 & 72.45 & 44.47 & 40.73 & 33.19 & 12.60 \\
\bottomrule
\end{tabular}
\end{table*}

\section{Qualitative analysis}
\label{suppsec:qualitative_analysis}
\begin{figure}[t]
  \centering
  \includegraphics[width=1.0\linewidth]{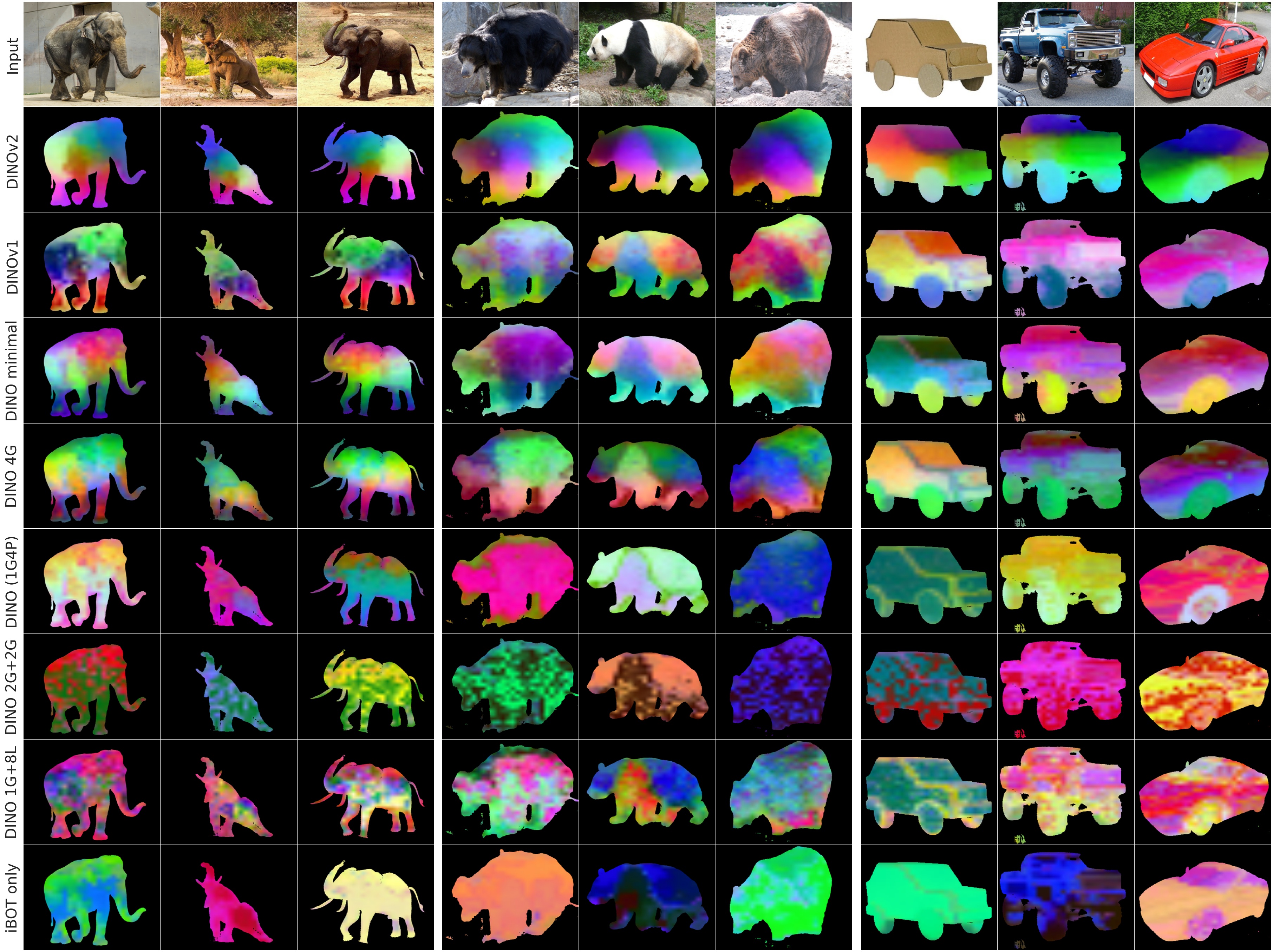}
  \caption{%
\textbf{PCA visualization of patch features under different component interventions.}
Each row shows RGB maps of the first three principal components (PC1--3) of ViT-L/16
patch features, fitted jointly over foreground tokens per category
(elephant, bear, vehicle; backgrounds masked to black). 
% From top to bottom, each row corresponds to rows \textsc{a}\oldstylenums{1}, \textsc{a}\oldstylenums{2}, \textsc{b}\oldstylenums{1}, \textsc{c}\oldstylenums{3}, \textsc{c}\oldstylenums{4}, \textsc{c}\oldstylenums{2}, \textsc{c}\oldstylenums{1} in~\cref{tab:main}, respectively.
\textit{DINO minimal} (2 global crops, same image) already produces spatially coherent,
part-consistent colorings across instances, demonstrating that global same-instance crops
alone suffice for semantic feature emergence.
More global crops from the same image (\textit{DINO 4G}) further improve consistency,
while cross-instance views (\textit{DINO 2G+2G}) and replacing global crops with local
ones (\textit{DINO 1G+8L}: 1 global + 8 local) both degrade structure severely.
Training with iBOT alone is similarly ineffective, confirming that the DINO
self-distillation loss is the operative signal.
Standard \textit{DINOv1} (2G+8L, DINO loss only: PCK@.1 = 32.0) closely matches
\textit{DINO minimal} (31.2), confirming that local crops add only marginal benefit.
\textit{DINOv2} (same 2G+8L recipe plus iBOT: PCK@.1 = 42.5) demonstrates that
the iBOT patch loss, not the local crops, drives the large gain over \textit{DINO minimal}.
Quantitative results in~\cref{tab:main} corroborate these observations.
  }
  \label{fig:supp_qual}
\end{figure}

We compare the spatial structure of ViT-L/16 patch features across a controlled set of
training objectives by projecting each image's patch tokens onto the first three
principal components (PC1--3) and rendering them as an RGB map
(Figure~\ref{fig:supp_qual}).
PCA is fitted jointly over foreground tokens within each category, so that consistent
colors across images reflect part correspondences rather than instance-specific appearance.

\paragraph{Same-instance global crops drive semantics.}
Already with two global crops of the same image (\textit{DINO minimal}), the PCA maps
show smooth, spatially coherent color gradients that are consistent across instances:
the trunk, body, and legs of different elephants receive the same relative hues,
and analogous structural regions of different vehicles are assigned similar colors.
These part-consistent groupings emerge from the self-distillation signal alone,
without local crops or patch-level objectives.
The maps of \textit{DINO 4G} are nearly indistinguishable from those of
\textit{DINO minimal} (colors may vary per row, but correspondence remains consistent), consistent with the small quantitative gain
(PCK@.1: 31.2 $\to$ 32.9).
When views share the same spatial crop and differ only photometrically
(\textit{DINO 1G4P}: 1 global crop, 4 color-jittered variants), the smooth
gradients give way to patchier, less consistent colorings; semantic part groupings
are visibly weaker, indicating that geometric diversity across views is a
necessary condition for structure to emerge.

\paragraph{Views must come from the same image instance.}
Replacing same-image crops with views drawn from two different image instances
(\textit{DINO 2G+2G}) produces visually fragmented, mottled maps: the smooth
gradients are replaced by a speckled, texture-like pattern with no stable part
groupings across instances.
Despite retaining global crops with geometric diversity, the network cannot form
part-consistent representations when views depict the semantics of different image instances;
the self-distillation target is corrupted at its source.

\paragraph{Local crops do not drive semantic structure.}
The PCA maps of \textit{DINOv1} (2G+8L, DINO loss) are consistent with those of \textit{DINO minimal} (2G, no local crops): the
same smooth gradients and part groupings appear in both rows, with differences
difficult to perceive at this scale, consistent with a negligible quantitative
change (PCK@.1: 31.2 $\to$ 32.0).
Removing global context and keeping only local crops (\textit{DINO 1G+8L}:
1 global + 8 local) produces fragmented, incoherent colorings similar to the
cross-instance failure case, confirming that local crops alone cannot sustain
the spatial signal needed for semantic structure.
Training on the patch-level iBOT objective alone (\textit{iBOT only}) collapses
further: the maps are nearly uniform within each silhouette, with almost no spatial
color variation, indicating that the DINO self-distillation loss is the main operative signal and not the patch reconstruction.
\textit{DINOv2}, which adds the iBOT loss to the same 2G+8L recipe as
\textit{DINOv1}, produces the richest part structure visually; the gap between
them traces directly to the iBOT objective, while the foundational semantic
organisation is already established in \textit{DINO minimal}.

\section{Bootstrap confidence intervals for correlation analysis}
\label{suppsec:bootstrap_ci}
\begin{figure}[t]
    \centering
    \includegraphics[width=0.78\linewidth]{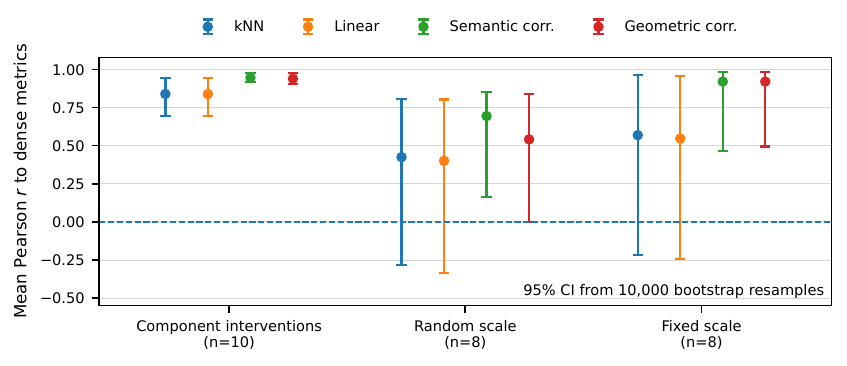}
    \caption{
    \textbf{Bootstrap confidence intervals for correlation analysis.}
    Points show the mean Pearson correlation between each proxy metric and dense
    downstream metrics across ablation variants; error bars indicate 95\%
    confidence intervals estimated from 10,000 non-parametric bootstrap resamples.
    Depth RMSE is sign-flipped before computing correlations so that higher
    values consistently indicate better downstream performance.
    }
    \label{fig:bootstrap-correlation-ci}
\end{figure}

\Cref{fig:bootstrap-correlation-ci} shows 95\% bootstrap confidence intervals (10,000 non-parametric resamples) for the Pearson correlations between each proxy metric and each dense downstream task, across all three experimental settings (component interventions, random-scale sweep, fixed-scale sweep).
Depth RMSE is sign-flipped before computing correlations so that higher values consistently indicate better downstream performance.
The intervals confirm that semantic and geometric correspondence metrics achieve higher mean correlation with dense tasks than kNN or linear probe across all settings, with the gap most pronounced in the fixed-scale sweep.

% \section{Additional analyses}
% \label{suppsec:more_analyses}
% \input{tables/tab_extra_supp}

%%%%%%%%%%%%%%%%%%%%%%%%%%%%%%%%%%%%%%%%%%%%%%%%%%%%%%%%%%%%

\end{document}